# DrMAD: Distilling Reverse-Mode Automatic Differentiation for Optimizing Hyperparameters of Deep Neural Networks

Jie Fu∗  Hongyin Luo^  Jiashi Feng∗  Kian Hsiang Low∗  Tat-Seng Chua∗

∗ National University of Singapore  ^ Tsinghua University

## Abstract

The performance of deep neural networks is well-known to be sensitive to the setting of their hyperparameters. Recent advances in reverse-mode automatic differentiation allow for optimizing hyperparameters with gradients. The standard way of computing these gradients involves a forward and backward pass of computations. However, the backward pass usually needs to consume unaffordable memory to store all the intermediate variables to *exactly* reverse the forward training procedure. In this work we propose a simple but effective method, DrMAD, to distill the knowledge of the forward pass into a shortcut path, through which we *approximately* reverse the training trajectory. Experiments on two image benchmark datasets show that DrMAD is at least 45 times faster and consumes 100 times less memory compared to state-of-the-art methods for optimizing hyperparameters with minimal compromise to its effectiveness. To the best of our knowledge, DrMAD is the first research attempt to make it practical to automatically tune *thousands* of hyperparameters of deep neural networks. The code can be downloaded from https://github.com/bigaidream-projects/drmad

## 1 Introduction

Modern machine learning algorithms are rarely hyperparameter-free. Hyperparameters, such as learning rate or $L2$-norm penalties are important for training deep models. Many works [Bergstra *et al.*2011, Shahriari *et al.*2016, Snoek *et al.*2012] have shown that the performance of large-sized deep models is sensitive to the setting of their hyperparameters. So, tuning hyperparameters is now recognized as a crucial step in the process of applying machine learning algorithms to achieve best performance and drive industrial applications [Shahriari *et al.*2016]. For decades, the de-facto standard for hyperparameter tuning in machine learning has been a simple grid search [Shahriari *et al.*2016]. Recently, it has been shown that optimizing hyperparameter in a principled and automatic way can reach or surpass human expert-level hyperparameter settings for deep neural networks in a variety of benchmark datasets [Shahriari *et al.*2016, Snoek *et al.*2012].

A common choice for hyperparameter optimization is gradient-free Bayesian optimization [Wang *et al.*2013]. Bayesian optimization builds a probability model to describe the distribution of validation loss conditioned on specific hyperparameters, which is obtained by multiple observations over the pairs of hyperparameter and validation loss. This probability model is then used to optimize the validation loss after complete training of the model's elementary[1] parameters. Although those techniques have been shown to achieve good performance with a variety of models on benchmark datasets [Shahriari *et al.*2016], they can hardly scale up to handle more than 20 hyperparameters [Maclaurin *et al.*2015, Shahriari *et al.*2016]. Here we mean *effective* hyperparameters. It has been shown in [Wang *et al.*2013] that Bayesian optimization can handle high-dimensional inputs only if the number of effective hyperparameters is small. Due to this inability, hyperparameters are often considered nuisances, encouraging researchers to develop machine learning algorithms with fewer of them. We argue that being able to richly hyperparameterize our models is more than a pedantic trick. For example we can set a separate $L2$-norm penalty for each layer,which has been shown to improve the performance of deep models on several benchmark datasets [Snoek *et al.*2012].

On the other hand, automatic differentiation (AD), as a mechanical transformation of an objective function, can calculate gradients with respect to hyperparameters (thus called hypergradients) accurately [Baydin *et al.*2015, Maclaurin *et al.*2015]. Although hypergradients enable us to optimize thousands of hyperparameters, all the prior attempts [Bengio2000, Baydin *et al.*2015, Maclaurin *et al.*2015] insist on *exactly* tracing the training trajectory backwards, which is impossible for real-world data and deep models from a memory perspective. Suppose we are to train a neural network on MNIST dataset, the iteration number is 200,000, and every elementary parameter vector takes up 0.1 GB. In order to trace the training trajectory backwards, a naïve solution has to store all the intermediate variables (e.g. weights) at every iteration, thus costing memory up to $200,000 \times 0.1$ GB = 20 TB. The improved method proposed in [Maclaurin *et al.*2015] needs at least 100 GB memory even for this small-scale dataset.

---
[1]Following the convention in [Maclaurin *et al.*2015], we use *elementary* to unambiguously denote the *traditional* parameters updated by back-propagation, e.g. weights and biases in a neural network.

The high memory cost stems from pursuing the exact reverse of the training procedure. In this work, we propose Distilling Reverse-Mode Automatic Differentiation (DrMAD), to reduce the memory cost and make the hyperparameter optimization feasible in practice. Compared with exact reverse methods, DrMAD chooses to reverse training dynamics in an *approximate* manner. Doing so allows us to reduce the memory consumption of tuning hyperparameters by a factor of 100,000 at least. On the MNIST dataset, our method only needs 0.2 GB memory, thus enabling the use of modern GPUs. More importantly, the memory consumption is independent of the problem size as long as the deep model has converged. Our convergence requirement is reasonable, as it has been demonstrated that modern deep neural networks are relatively easy to converge in practice [Choromanska *et al.*2014, Goodfellow and Vinyals2014]. In addition, DrMAD only incurs negligible performance drop. Section 2 and Section 3 describe this problem and our solution in detail respectively, which is the main technical contribution of this paper.

In short, we make the following contributions in this work:

- We propose an algorithm that approximately reverses stochastic gradient descent to compute gradients w.r.t. hyperparameters. Our method can reduce memory consumption significantly without sacrificing its effectiveness, compared with state-of-the-art methods based on automatic differentiation.

- We propose a hyperparameter server framework to solve distributed hyperparameter optimization problems.

## 2 Background and Related Work

We review the general framework of automatic hyperparameter tuning and previous work on automatic differentiation for hyperparameters of machine learning models.

A learning algorithm $\mathcal{A}_{w,\lambda}$ is described by a vector of $m$ elementary parameters $\boldsymbol{w} = (w_1, ..., w_m) \in \boldsymbol{W}$, where $\boldsymbol{W} = W_1 \times ... \times W_m$ define the parameter space, and a vector of $n$ hyperparameters $\boldsymbol{\lambda} = (\lambda_1, ..., \lambda_n) \in \boldsymbol{\Lambda}$, where $\boldsymbol{\Lambda} = \Lambda_1 \times ... \times \Lambda_n$ define the hyperparameter space. We further use $l_{train} = \mathcal{L}(\mathcal{A}_{w,\lambda}, \mathrm{X}_{train})$ to denote the training loss, and $l_{valid} = \mathcal{L}(\mathcal{A}_{w,\lambda}, \mathrm{X}_{valid}; \mathrm{X}_{train})$ to denote the validation loss that $\mathcal{A}_{w,\lambda}$ achieves on validation data $\mathrm{X}_{valid}$ when trained on training data $\mathrm{X}_{train}$. An automatic hyperparameter tuning algorithm then tries to find $\boldsymbol{\lambda} \in \boldsymbol{\Lambda}$ that minimizes $l_{valid}$ in an efficient and principled way.

To make the definitions more concrete and concise, we further denote the training objective function as: $l_{train} = \mathcal{L}(\boldsymbol{w}|\boldsymbol{\lambda}, \mathrm{X}_{train}) = C(\boldsymbol{w}|\boldsymbol{\lambda}, \mathrm{X}_{train}) + P(\boldsymbol{w}, \boldsymbol{\lambda}) = C_{train} + P(\boldsymbol{w}, \boldsymbol{\lambda})$, where $C(\cdot)$ is the cost function on either training (denoted by $C_{train}$) or validation (denoted by $C_{valid}$) data, and $P(\cdot)$ is the penalty term.

### 2.1 Automatic differentiation (AD)

Most training of machine learning models is driven by the evaluation of derivatives, which can be handled by automatic differentiation (AD). AD systematically applies the chain rule of calculus at the elementary operator level [Griewank and Walther2008]. It also guarantees the accuracy of evaluation of derivatives with a small constant factor of computational overhead and ideal asymptotic efficiency [Baydin *et al.*2015].

| Forward Pass | | |
|---|---|---|
| $v_{-1}$ | $= x_1$ | $= 3$ |
| $v_0$ | $= x_2$ | $= 6$ |
| $v_1$ | $= \ln(v_0)$ | $= \ln(6)$ |
| $v_2$ | $= v_{-1}^2$ | $= 3^2$ |
| $v_3$ | $= \cos(v_{-1})$ | $\cos(3)$ |
| $v_4$ | $= v_1 + v_2$ | $= 1.79 + 9$ |
| $v_5$ | $= v_4 + v_3$ | $= 10.79 - 0.98$ |
| $y$ | $= v_5$ | $= 9.80$ |
| Backward Pass | | |
| $\bar{x}_1$ | $= \bar{v}_{-1}$ | $= 5.86$ |
| $\bar{x}_2$ | $= \bar{v}_0$ | $= 0.16$ |
| $\bar{v}_{-1}$ | $= \bar{v}_{-1} + \bar{v}_2 \frac{\partial v_2}{\partial v_{-1}}$ | $= 5.86$ |
| $\bar{v}_0$ | $= \bar{v}_1 \frac{\partial v_1}{\partial v_0}$ | $= 0.16$ |
| $\bar{v}_{-1}$ | $= \bar{v}_3 \frac{\partial v_3}{\partial v_{-1}}$ | $= -0.14$ |
| $\bar{v}_1$ | $= \bar{v}_4 \frac{\partial v_4}{\partial v_1}$ | $= 1$ |
| $\bar{v}_2$ | $= \bar{v}_4 \frac{\partial v_4}{\partial v_2}$ | $= 1$ |
| $\bar{v}_3$ | $= \bar{v}_5 \frac{\partial v_5}{\partial v_3}$ | $= 1$ |
| $\bar{v}_4$ | $= \bar{v}_5 \frac{\partial v_5}{\partial v_4}$ | $= 1$ |
| $\bar{v}_5$ | $= \bar{y}$ | $= 1$ |

Table 1: Reverse-mode automatic differentiation example, with $y = f(x_1, x_2) = \ln(x_2) + x_1^2 + \cos(x_1)$ at $(x_1, x_2) = (3, 6)$. Setting $\bar{y} = 1$, $\partial y / \partial x_1$ and $\partial y / \partial x_2$ are computed in one backward pass.

AD has two modes: forward and reverse [Griewank and Walther2008]. Here we only consider the reverse-mode automatic differentiation (RMAD) [Baydin *et al.*2015]. RMAD is a generalization of the back-propagation [Goodfellow *et al.*2015] used in the deep learning community. In fact, one of the most popular deep learning libraries, Theano [Bastien *et al.*2012], can be described as a limited version of RMAD and a heavily optimized version of symbolic differentiation [Baydin *et al.*2015]. RMAD allows the gradient of a scalar loss with respect to its parameters to be computed in a single backward pass after a forward pass [Baydin *et al.*2015]. Table 1 shows an example of RMAD for $y = f(x_1, x_2) = \ln(x_2) + x_1^2 + \cos(x_1)$, using the "three-part" notation in [Griewank and Walther2008], a trace of $f : \mathbb{R}^N \to \mathbb{R}^M$ is constructed from (a) $v_{i-N} = x_i$, i=1,...,N input variables, (b) $v_i, i = 1, ..., l$ working variables, and (c) $y_{M-i} = v_{l-i}$, $i = M - 1, ..., 0$ output variables.

### 2.2 Gradient-based methods for hyperparameters

In this paper we focus on studying the tuning of continuous hyperparameters. However, we can still hyperparameterize certain discrete designs using our proposed method, which will be shown in Section 4.3. We consider stochastic gradient descent (SGD), as it is widely used to optimize large-sized neural networks [Goodfellow *et al.*2015].

The elementary parameters updating formula is: $\boldsymbol{w}_{t+1} = \boldsymbol{w}_t + \eta_{\boldsymbol{w}} \nabla_{\boldsymbol{w}} \mathcal{L}(\boldsymbol{w}_t | \boldsymbol{\lambda}, \mathrm{X}_{train})$, where the subscript $t$ denotes the count of iteration (i.e. one forward and backward pass over one mini-batch), and $\eta_{\boldsymbol{w}}$ is the learning rate for elementary parameters.

The gradients of hyperparameters (hypergradients) are computed on the validation data $\mathrm{X}_{valid}$ without considering the penalty term [Foo *et al.*2008, Maclaurin *et al.*2015, Luketina *et al.*2015]:

$$\nabla_{\boldsymbol{\lambda}} C_{valid} = \nabla_{\boldsymbol{w}} C_{valid} \frac{\partial \boldsymbol{w}_t}{\partial \boldsymbol{\lambda}} = \nabla_{\boldsymbol{w}} C_{valid} \frac{\partial^2 l_{train}}{\partial \boldsymbol{\lambda} \partial \boldsymbol{w}}, \quad (1)$$

where $C_{valid} = C(\boldsymbol{w} | \mathrm{X}_{valid})$ is the validation cost.

The hyperparameters are updated at every iteration in [Luketina *et al.*2015, Foo *et al.*2008]. In [Foo *et al.*2008], given the elementary optimization has converged, the hyperparameters are updated as:

$$\boldsymbol{\lambda}_{t+1} = \boldsymbol{\lambda}_t + \eta_{\boldsymbol{\lambda}} \nabla_{\boldsymbol{w}} C_{valid} (\nabla_{\boldsymbol{w}}^2 l_{train})^{-1} \frac{\partial^2 l_{train}}{\partial \boldsymbol{\lambda} \partial \boldsymbol{w}}, \quad (2)$$

where $\eta_{\boldsymbol{\lambda}}$ is the learning rate for hyperparameters. The authors in [Luketina *et al.*2015] propose to update hyperparameters by simply approximating the Hessian in Eq. 2 as $\nabla_{\boldsymbol{w}}^2 l_{train} = I$:

$$\boldsymbol{\lambda}_{t+1} = \boldsymbol{\lambda}_t + \eta_{\boldsymbol{\lambda}} \nabla_{\boldsymbol{w}} C_{valid} \frac{\partial^2 l_{train}}{\partial \boldsymbol{\lambda} \partial \boldsymbol{w}}. \quad (3)$$

However, updating hyperparameters at every iteration would result in unstable hypergradients. Because this approach only considers the influence of the regularization hyperparameters on the *current* elementary parameter update, it can hardly scale up to handle more than 20 hyperparameters as shown in [Luketina *et al.*2015].

In this paper, we adopt RMAD for computing hypergradients, like [Maclaurin *et al.*2015, Bengio2000, Domke2012] by taking into account the effects of the hyperparameters on the entire learning trajectory. Specifically, different from Eq. 1 that only considers $\frac{\partial \boldsymbol{w}_t}{\partial \boldsymbol{\lambda}}$, DrMAD considers the term $\frac{\partial \boldsymbol{w}_T}{\partial \boldsymbol{\lambda}}$ (here $T$ represents the final iteration number till convergence): $\boldsymbol{w}_T = \sum_{0 < t < T} \triangle \boldsymbol{w}_{t,t+1}(\boldsymbol{w}_t(\boldsymbol{\lambda}_k), \boldsymbol{\lambda}_k, \mathrm{X}_t, \eta_{\boldsymbol{\lambda}}) + \boldsymbol{w}_0$, where the subscript $k$ in $\boldsymbol{\lambda}_k$ stands for the counter of meta-iterations used for hyperparameter optimization (i.e. the number of entire training of elementary parameters), $\mathrm{X}_t$ is the mini-batch of training data used in iteration $t$, and $\boldsymbol{w}_0$ is the initial parameter vector. Update of hyperparameters in this paper and also [Maclaurin *et al.*2015, Bengio2000, Domke2012] is:

$$\boldsymbol{\lambda}_{k+1} = \boldsymbol{\lambda}_k + \eta_{\boldsymbol{\lambda}} \nabla_{\boldsymbol{w}} C_{valid} \frac{\partial \boldsymbol{w}_T}{\partial \boldsymbol{\lambda}}. \quad (4)$$

Unfortunately, RMAD requires all the intermediate variables obtained in the forward pass should be maintained in memory for the backward pass [Griewank and Walther2008]. Conventional RMAD with exact arithmetic stores the entire training trajectory $\{\boldsymbol{w}_0, ..., \boldsymbol{w}_T\}$ in memory, which is totally impractical for even small-sized tasks.

An "information buffer" method is proposed in [Maclaurin *et al.*2015] to recompute the learning trajectory on the fly during the backward pass of RMAD rather than storing it in memory by making use of the SGD momentum mechanism. Certain amount of auxiliary bits as information buffer, which depend on the specific learning dynamics, to handle finite precision arithmetic is needed in this case. Although the proposed method in [Maclaurin *et al.*2015] is shown to be able to tune thousands of hyperparameters and reduce the memory consumption by a factor of 200 in the most ideal setting, it needs 8 hours to run 10 epochs on a subset (10,000 training samples) of MNIST dataset. Furthermore, its memory and computational requirements grow without bound as the problem size increases, which makes it impossible to run on modern GPUs. In contrast, we will present in the next section how to approxiate RMAD to reduce memory consumption dramatically, thus enabling it to leverage the power of GPUs in principle.

## 3 Approximating RMAD with a Shortcut

In this paper, we raise a crucial yet rarely investigated question: do we really need to *exactly* trace the whole training procedure backwards, starting from the trained parameter values and working back to the initial random parameters? If all that we care about are good enough hyperparameters given limited resources and time budget, the answer might be no. Driven by this question, we demonstrate that a shortcut can be established by distilling the knowledge from the forward pass of RMAD with minimal compromise to its solution quality.

### 3.1 Distilling knowledge from the forward pass

Trying to trace backwards exactly can be wasteful as this approach does not take into account the highly structured nature of training dynamics of deep models. One intuition motivating the investigation of this paper is the observation [Goodfellow and Vinyals2014] that if we knew the direction defined by the final learned weights after convergence, a single coarse line search is sufficient to visualize the training dynamics of a neural network. In other words, there exists a linear subspace in which the training of a neural network rarely encounters any significant difficulties. This observation is consistent with other recent empirical and theoretical work demonstrating that modern deep neural networks are relatively easy to optimize [Choromanska *et al.*2014, Montufar *et al.*2014].

It is also well-known that when doing elementary parameter optimization, second-order methods are more efficient with fewer iterations compared to the naïve gradient descent, but they cannot be easily applied to high-dimensional models due to heavy computations [Raiko *et al.*2012]. In practice, it is usually approximated by a diagonal or block-diagonal approximation [Schaul and LeCun2013]. Inspired by these approximation techniques, we make an aggressive approximation for RMAD here – discard all the intermediate variables altogether. In other words, we choose a shortcut, which simply approximates the forward pass learning history used in Eq. 4 as a series of parameter vectors $\boldsymbol{w} = (1-\beta)\boldsymbol{w}_0 + \beta \boldsymbol{w}_T$ for varying values of $0 < \beta < 1$, which can be generated on the fly almost without storing anything. Figure 1 shows the contour for the backward passes used by DrMAD and RMAD respectively.

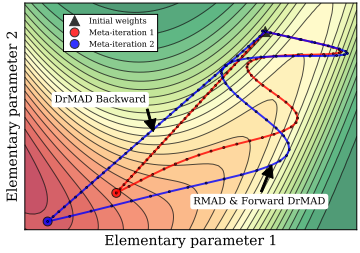

Figure 1: Comparisons on the paths of hyperparameter optimization by DrMAD and RMAD. RMAD uses the same path for its forward and backward passes. While the backward pass of DrMAD follws a shortcut, which is established by distilling the knowledge from its forward pass.

More concretely, DrMAD works by first obtaining the final trained elementary parameter values using SGD algorithms. Algorithm 1 demonstrates the procedure formally. Here we could use any SGD variant and do not put constraints on the momentum term. In contrast, the previous most similar work [Maclaurin *et al.*2015] is highly dependent on the momentum setting in order to save memory during the backward pass of RMAD.

Algorithm 2 shows how to compute the gradients of hyperparameters by DrMAD, where in step 4 we approximate the learning dynamics. These hypergradients are used to update hyperparameters using Eq. 4. In DrMAD, the approximated intermediate elementary parameters are independent of each other. Whereas in RMAD proposed in [Maclaurin *et al.*2015] each elementary parameter relies on the previous one implicitly, and thus a "information buffer" is needed. Since that extra buffer is dataset and model dependent, RMAD's memory consumption is growing without bound.

In addition to reduction in memory, compared to [Maclaurin *et al.*2015], Algorithm 2 also reduces the computational operations as a byproduct, because it does not need to recompute the elementary parameters exactly.

---

**Algorithm 1** Stochastic gradient descent (SGD).

1: **inputs:** initial parameters $w_0$, learning rate $\alpha$, weight decay $\gamma$, hyperparameters $\lambda$, loss function $l_{train}$, iteration number $T$.
2: initialize $v_1 \leftarrow 0$
3: **for** $t = 1$ to $T - 1$ **do**
4: $\quad g_t \leftarrow \nabla_w l_{train}$ //evaluate gradient
5: $\quad v_{t+1} \leftarrow \gamma v_t - (1 - \gamma) g_t$
6: $\quad w_{t+1} \leftarrow w_t + \alpha_t v_t$ //update position
7: **end for**
8: **output:** trained parameters $w_T$

---

### 3.2 Discussions on DrMAD

In fact, DrMAD can be seen as an operation of knowledge distillation. As explained in [Hinton *et al.*2014], the knowledge acquired by an ensemble of large-sized models as a "teacher" can be *distilled* into a single small model, which is called the "student". For example, one can construct a teacher model by training an ensemble of 5 deep neural networks with 10 layers. Then, a student model with 5-layer can be constructed to achieve similar performance as the teacher. However, it requires much less parameters as it only approximates the teacher's behavior rather than learning from scratch.

Considering our situation, the most valuable knowledge is the initial random and final trained weight vectors once the training process of a deep model has converged. Therefore, the forward pass of SGD training provides the final weight vector after convergence, which is defined as a teacher; the reverse pass given by shortcut is a student here.

---

**Algorithm 2** Distilling reverse-mode automatic differentiation (DrMAD) of SGD.

1: **inputs:** initial parameter $w_0$, learned parameter $w_T$, training loss $l_{train}$, validation loss $l_{valid}$, weight decay $\gamma$, learning rate $\alpha$
2: initialize $d\lambda \leftarrow 0$, $dw \leftarrow \nabla_w l_{valid}$, $\beta_{T-1} \leftarrow 1 - \frac{1}{T}$
3: **for** $t = T - 1$ to $1$ **do**
4: $\quad w_{t-1} \leftarrow (1 - \beta_t) w_0 + \beta_t w_T$
5: $\quad \beta_{t-1} \leftarrow \beta_t - \frac{1}{T}$
6: $\quad dv \leftarrow dv + \alpha dw$
7: $\quad d\lambda \leftarrow d\lambda - (1 - \gamma) dv \nabla_\lambda \nabla_w l_{train}$
8: **end for**
9: **output:** gradient of $l_{valid}$ w.r.t. $\lambda$

---

Obviously, obtaining derivatives from the shortcut may never reveal more information about hyperparameters than calculating derivatives from the exact trajectories. However, when memory and computational costs are taken into account, the shortcut may convey more information per unit cost. In fact, our approach explicitly separates the optimization of the elementary parameters and hyperparameters, which serves as a trade-off between accuracy and computational expense. In the experimental part, we will show that the accuracy performance of DrMAD is slightly worse than the RMAD with exact arithmetic. Nonetheless, this separation gives rise to several positive implications. For example, we can use distributed deep learning libraries, such as CNTK[2], to speed up the forward pass, because currently none of the automatic differentiation libraries, such as Autograd[3], support multiple GPUs. DrMAD is also orthogonal to the choices associated with the internal elementary parameter optimization; it can work alongside other recent advances in neural network training without modification.

### 3.3 Hyperparameter server

With current generation hardware such as large computer clusters with GPUs, the optimal allocation of computing cycles should include more hyperparameter exploration than has been typical in the past [Bergstra *et al.*2011]. It is thus desirable to parallelize hyperparameter optimization processes. Motivated by the parameter server approach to distributed elementary parameter optimization [Li *et al.*2013], we propose a *hyperparameter server* framework for distributed hyperparameter optimization using hypergradients. In parameter server [Li *et al.*2013], computational nodes are partitioned into clients and servers, and communication between nodes is asynchronous. In our hyperparameter server framework, there is only one server and several clients, and the communication between them is synchronous. Each client is in charge of training a model replica with the same hyperparameters on a subset of data, and the server maintain the globally shared hyperparameters. At every meta-iteration, we accumulate the hypergradients from the clients and average them. The averaged hypergradients then are used to update the global hyperparameters maintained by the server. The overall architecture is shown in Figure 2.

---

[2]https://github.com/Microsoft/CNTK
[3]https://github.com/HIPS/autograd

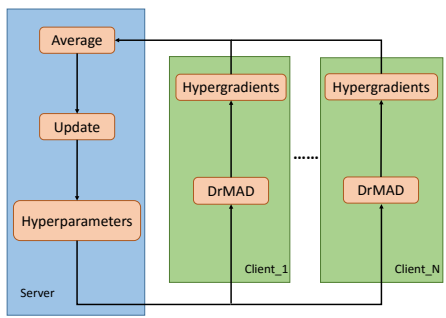

Figure 2: Hyperparameter server architecture.

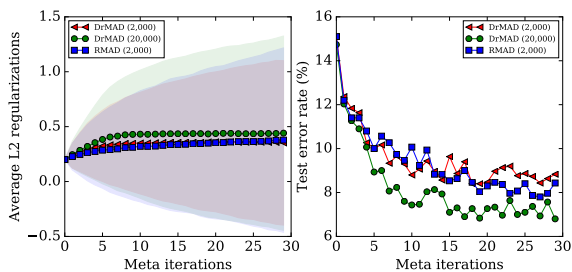

Figure 3: *Left:* The average values of hyperparameters with variances obtained by DrMAD (2,000 iterations), DrMAD (20,000 iterations) and RMAD (2,000 iterations), respectively. *Right:* The test error rates obtained by DrMAD (2,000 iterations), DrMAD (20,000 iterations) and RMAD (2,000 iterations), respectively.

## 4 Experiments

In this section, we empirically demonstrate how DrMAD offers high memory and computational efficiency yet achieving comparable predictive performance as computationally expensive RMAD. Note that here we do not strive for state-of-the-art performance on benchmark datasets, but focus on showcasing the merits of DrMAD by comparing it with RMAD. Therefore, even though we use image data in the experiments, we do not use or compare to any image-specific processing such as convolutional networks. Note that even the best results achieved by RMAD [Maclaurin *et al.*2015] are not comparable with those obtained with for instance convolutional networks. All the experiments are done on a Microsoft Azure G5 server with 32 CPU cores and 448 GB memory.

### 4.1 Optimizing continuous regularization hyperparameters

Although DrMAD could work in principle for many different types of continuous hyperparameters, we focus on tuning hyperparameters for regularization here. Following the settings in [Maclaurin *et al.*2015], we evaluate DrMAD for optimizing continuous regularization hyperparameters on a subset of MNIST dataset (10,000 for training, 3,000 for validation, and 3,000 for testing) using a multilayer perceptron (MLP) with *tanh* activation function. We do not include data shuffling and use a fixed set of random initial parameters for every meta-iteration. The MLP has 4 layer, containing 784, 50, 50, and 50 neurons respectively. Each neuron has its own $L$2-norm penalty on its parameter and thus we are going to optimize 934 hyperparameters in total. The learning rate for optimizing elementary parameters is fixed as 0.05, the number of elementary iterations is set as 2000, the learning rate for hyperparameters is 0.07, the mini-batch size is 50, and the number of meta-iterations is 30. The adopted data pre-processing operations only includes centering each feature. The code of DrMAD is modified from a public toolbox for hypergradient calculation[4]. We use the original code from [Maclaurin *et al.*2015] for experiments on RMAD.

Figure 3 shows the curve of averaged hyperparameter values w.r.t. optimization iterations and the corresponding test error from DrMAD and RMAD. The running time of DrMAD and RMAD are 16 minutes and 717 minutes, respectively. One can observe that the average hyperparameter values and test error rate curves of DrMAD are close to RMAD on MNIST dataset while DrMAD only costs 2% computation time of RMAD.

In contrast to [Maclaurin *et al.*2015], we set the mini-batch size to a small number (v.s. 300 in [Maclaurin *et al.*2015]) and the number of elementary iterations to a large one (v.s. 200 elementary iterations in [Maclaurin *et al.*2015]) on purpose, because these settings result in highly zigzag and long learning trajectories. The shaded areas in Figure 3 (left) represent the variances of individual hyperparameters. We can see that the variances are quite high, which implies that diverse hyperparameter values might be beneficial to the predictive performance. Another observation is that the tuned hyperparameters seem to prefer positive values. Figure 3 (right) shows that DrMAD can provide similar performance of RMAD. Also note that DrMAD consumes 100 times less memory than RMAD does.

To demonstrate that DrMAD can handle even longer learning trajectories, we increase the number of elementary iterations to 20,000. But due to the time budget, we do not compare it with RMAD, which is at least 45 times slower. Figure 3 (left) shows that for DrMAD with 20,000 iterations, the mean of hyperparameters tends to be larger compared to the ones obtained using 2,000 iterations. Also, the variances of the hyperparameters obtained by DrMAD with 20,000 iterations are towards larger positive values. One possible explanation could be that taking more iterations increases the risk of overfitting, and thus heavier penalties are needed. We can observe in Figure 3 (right) that DrMAD with 20,000 iterations achieves better performance compared to DrMAD and RMAD with only 2,000 iterations.

---
[4]https://github.com/HIPS/hypergrad

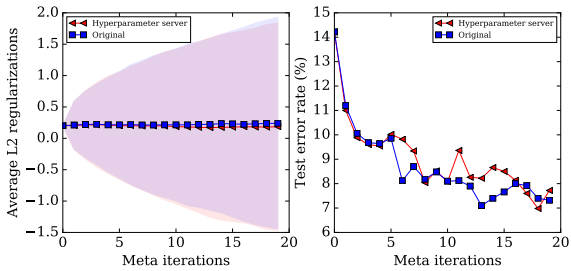

Figure 4: *Left:* The average values of hyperparameters with variances obtained by the hyperparameter server version of DrMAD and the original DrMAD, respectively. *Right:* Test error rates obtained by the hyperparameter server version of DrMAD and the original DrMAD, respectively.

## 4.2 Hyperparameter server

We apply the hyperparameter server framework to a subset of MNIST dataset. Specifically, the number of total training datapoints is 30,000, and the numbers of both validation and test data are 5,000. For the hyperparameter server based DrMAD, we have 5 clients, each being trained with 12,000 training samples. The number of meta-iterations is 20. All the other settings are the same as our previous experiments.

Figure 4 (right) shows the effectiveness of our proposed hyperparameter server framework on MNIST dataset. For the hyperparameter server, the test errors are estimated by a model trained with 30,000 datapoints and the average hyperparameters from clients. We can observe that in Figure 4 (left), the evolution of the average values and the variances of L2 penalties obtained by the hyperparameter server looks similar to that of the original DrMAD. Figure 4 (right) shows that DrMAD with the hyperparameter server framework can approach the performance of the original DrMAD. Overall, it seems that diversity of hyperparameters is beneficial to the predictive performance of deep neural networks on MNIST dataset.

We can see from Figure 4 (left) that the variance of hyperparameters seems unbounded. One might be tempted to further increase the number of meta-iterations to make the variance even larger. But Figure 4 (right) indicates that with the increase in variance, the predictive performance increases initially and then stops changing dramatically. Furthermore, according to our experiments and the observations in [Maclaurin *et al.*2015], hypergradients become unstable after certain number of meta-iterations.

The computations of hypergradients are dependent on their hyperparameters through thousands of iterations of SGD. Furthermore, within each iteration of SGD, it involves forward- and then back-propagations through a deep neural network. Overall, the stacking of all the above operations would result in vanishing gradient problems [Glorot and Bengio2010]. It should be noted that both DrMAD and RMAD have this problem.

We would argue that this instability may not be a serious problem in practice, because due to the limited computing and time budget, running too many meta-iterations (e.g. 50) is not reasonable anyway.

## 4.3 Learning continuously hyperparameterized architectures

Popular deep learning architectures, such as convolutional neural networks, can be obtained by forcing particular weights to be zero and tying particular pairs of weights together with hard constraints. It has been shown in [Maclaurin *et al.*2015] that the learning of softened architectural constraints can be seen as a multitask learning problem using a MLP with 2 hidden layers. They learn a penalty for each alphabet pair, separately for each layer in neural networks. Specifically, the hypergradients are computed by a pairwise quadratic penalty on the hyperparameters, $\boldsymbol{\lambda}^T A \boldsymbol{\lambda}$, where $A$ is described by three $10 \times 10$ matrices, each matrix for a particular layer, and $\boldsymbol{\lambda}$ is the hyperparameter vector. They also use the Omniglot dataset [Lake *et al.*2015]. The dataset consists of 10 alphabets with up to 55 characters in each alphabet but only 15 examples of each character. Each character is represented by a $28 \times 28$ pixel greyscale image. More details about the dataset are presented in [Lake *et al.*2015].

Here, we reproduce the above experiment with the same settings in [Maclaurin *et al.*2015], but use DrMAD instead. Our experiments show that the training and test error obtained by DrMAD are 0.42 and 1.13 respectively, whereas RMAD's training and test error are 0.60 and 1.13 respectively. If we increase the number of elementary iterations from 50 to 2,000, the training and test error of DrMAD are 0.52 and 1.10 respectively.

## 5 Conclusion and Future Work

In this paper, we proposed a highly memory efficient hyperparameter optimization method – distilling reverse-mode automatic differentiation (DrMAD) to optimize continuous hyperparameters in deep neural networks. We demonstrated how DrMAD is able to optimize validation loss w.r.t. thousands of hyperparameters in practice, which was previously impossible due to its unreasonably large memory consumption. We demonstrated its effectiveness and efficiency on two benchmark datasets.

While DrMAD gives comparable results on small image datasets, it is not clear if this approach will work for larger datasets. Since DrMAD makes it possible for the first time to run on GPUs, rewriting DrMAD to train models on larger datasets with GPUs would be our future work.

### Acknowledgments
Jie Fu thanks Microsoft Azure for Research for providing the computational resources. This work is also supported by NUS-Tsinghua Extreme Search (NExT) project through the National Research Foundation, Singapore.